\documentclass[runningheads]{llncs}
\usepackage[T1]{fontenc}
\usepackage{amssymb}
\usepackage{graphicx}
\usepackage{multirow}
\usepackage{booktabs}
\usepackage{graphicx}
\usepackage{bbm}
\usepackage{amsmath}
\usepackage{engord}
\usepackage{color}
\usepackage[colorlinks,
linkcolor=red,       
anchorcolor=blue,  
citecolor=blue,        
]{hyperref}
\begin{document}
\title{Cross-supervised Dual Classifiers for Semi-supervised Medical Image Segmentation}
%
%
\author{Zhenxi Zhang\inst{1} \and
	Ran Ran\inst{2,3} \and
	Chunna Tian\inst{1} \and Heng Zhou \inst{1} \and Fan Yang\inst{4} \and Xin Li\inst{4} \and Zhicheng Jiao\inst{5}}
%
%
\institute{Xidian University, 2 South Taibai Road, Shanxi Province \email{zxzhang\_5@stu.xidian.edu.cn} \and
	Cancer Center, The First Affiliated Hospital of Xi'an Jiaotong University \and Precision Medicine Center, The First Affiliated Hospital of Xi'an Jiaotong University \and AIQ, Abu Dhabi, United Arab Emirates
	 \and
	Department of Diagnostic Imaging, Warren Alpert Medical School of Brown University\\
	}
%

\authorrunning{Zhenxi Zhang et al.}
%

%
\maketitle              
\begin{abstract}
Semi-supervised medical image segmentation offers a promising solution for large-scale medical image analysis by significantly reducing annotation burden while achieving comparable performance. Employing this method exhibits a high degree of potential for optimizing the segmentation process and increasing its feasibility in clinical settings during translational investigations.
Recently, cross-supervised training based on different co-training sub-networks has become a common paradigm for this task, but the critical issues of sub-network disagreement and label-noise suppression require further attention and progress in cross-supervised training. In this paper, we propose a cross-supervised learning framework based on dual classifiers (DC-Net), including an evidential classifier and a vanilla classifier. The two classifiers exhibit complementary characteristics, enabling them to handle disagreement effectively and generate more robust and accurate pseudo-labels for unlabeled data. We also incorporate the uncertainty estimation from the evidential classifier into cross-supervised training to alleviate the negative effect of error supervision signal.
The extensive experiments on LA and Pancreas-CT dataset illustrate that DC-Net  outperforms other state-of-the-art methods for semi-supervised segmentation. The code will be released soon.
\keywords{Dual classifiers  \and Semi-supervised segmentation \and Co-training.}
\end{abstract}
\section{Introduction}
Medical image segmentation is a precondition of for computer-aided diagnosis and treatment in imaging-based research and clinical translation. Deep learning methods have achieved tremendous success in many medical image segmentation tasks under a fully-supervised training regime. However, due to the increasing data security concerns, particularly in multi-cohort settings, the pixel-level annotation is becoming less practical. Besides, accurate annotation is a time-consuming and arduous task that often requires specialized knowledge or expertise. Semi-supervised segmentation \cite{mittal2019semi}, which involves leveraging unlabeled data to reduce annotation workload without sacrificing model performance, has been a prevailing method.

Consistency regularization \cite{berthelot2019mixmatch,li2021dual} and pseudo-labeling \cite{yang2022st++} account for two mainstream semi-supervised medical image segmentation methods. The logic of consistency constraint is to encourage smooth predictions of the same data under different perturbations, e.g., using unlabeled data and implementing consistency regularization between different settings in terms of perturbed features \cite{ouali2020semi}, model initialization \cite{ke2020guided}, contextual information \cite{lai2021semi} or interpolation ways \cite{wu2021semi}. 

The pseudo-labeling approach involves using pre-trained models to assign labels to unlabeled data, and the model is then updated iteratively with these pseudo-labeled samples. Recently, cross supervision learning mechanism based on the co-training sub-networks further combine the benefits of the two approaches mentioned above. For example, 
Luo, et al \cite{luo2021semi} propose a dual-task consistency method to enforce the prediction consistency between the level set regression network and the segmentation network.
Wu, et al \cite{wu2021semi} introduce mutual consistency between two slightly different decoders to address the highly uncertain and easily mis-classified predictions in the ambiguous regions. 

Despite demonstrating potential, prior studies have failed to consider the essential factor of proper disagreement \cite{fan2022ucc} among sub-networks, which is critical for effective co-training and information extraction from unlabeled data. Inadequete disagreement may cause co-training to devolve into self-training that only employs labeled data, ultimately resulting in a consensus situation. Conversely, excessive disagreement could undermine the quality of the pseudo labels, leading to unreliable cross-supervision. 
To address this issue, we explore to ensure appropriate disagreement by implementing two separate sub-classifiers with the shared encoder. The first sub-classifier is a vanilla classifier that utilizes cross-entropy loss, a prevalent technique in pixel-wise classification known for its fast convergence and reliability, but it is prone to over-confidence. While, the second sub-classifier is an evidential non-negative classifier, which uses evidence theory \cite{sensoy2018evidential} to represent the degree of belief for each class and incorporate epistemic uncertainty into the pixel-wise classification process of segmentation tasks \cite{zou2022tbrats,li2022region}. 
The inclusion of two complementary sub-classifiers in the model design promotes diverse opinions and safeguards against excessive reliance on a single classifier.

Another critical concern is that the generation of noisy predictions in pseudo-labels typically happens in complex or blurred regions, leading to error accumulation that could degrade the model's overall performance.  To address this issue, several studies \cite{yu2019uncertainty,wang2021tripled} introduce Monte-Carlo dropout with high inference cost to estimate uncertainty and filter out the potential noisy pseudo labels. Nevertheless, the filter operation ignores some valuable
pseudo labels under the threshold. 
Additional efforts \cite{wu2021semi,luo2021semi} treat all pixel-wise predictions the same in the cross-supervised training process, which may introduce incorrect guidance  at the early training stage and tends to accumulate to the degree.  Our approach differs from existing solutions as we take the uncertainty resulting from evidential learning into account and integrate it into cross-supervised learning. Therefore, our model could adjust the weight of the pixel-wise cross-supervised loss, with the goal of reducing the adverse impact of unreliable pseudo labels. 

The main contributions of this paper are summarized as: (1) We propose a novel cross-supervised learning method based on dual classifiers (DC-Net) for semi-supervised medical image segmentation. (2) To our best knowledge, DC-Net is the first to introduce the evidential non-negative classifier to ensure the disagreement with vanilla classifier in cross-supervised training. (3) We design the evidential uncertainty-guided consistency learning method to alleviate the negative effect of the noise problem in pseudo labels. (4) We conduct extensive experiments on LA dataset \footnote[1]{ \url{http://atriaseg2018.cardiacatlas.org}} and Pancreas-CT dataset \footnote[2]{\url{ https://wiki.cancerimagingarchive.net/display/Public/Pancreas-CT}} to demonstrate that our method improves the semi-supervised segmentation performance and outperforms other state-of-the-art methods.

\section{Methodology}
Let $\mathcal{D}_l$ and $\mathcal{D}_u$ denote the labeled set and unlabeled set, respectively. The labeled data is denoted as $(x_l,y_l) \in \mathcal{D}_l$, and the unlabeled data is $x_u \in \mathcal{D}_u$. Our task is to improve the segmentation performance using $\mathcal{D}_u$ and $\mathcal{D}_l$ jointly based on the proposed DC-Net.
\subsection{Dual Classifiers}
The Softmax function in vanilla classifier can amplify the difference in classification scores. Thus, the difference in scores can be further magnified, which tends to make the classification results over-confident. While, the evidential classifier provides a way to quantify the prediction uncertainty, offering  a more nuanced and probabilistic perspective on pixel-wise classification. Therefore, the dual classifiers complement each other and provide useful supervision for each other.
\begin{figure}[t]
	\includegraphics[width=1\textwidth]{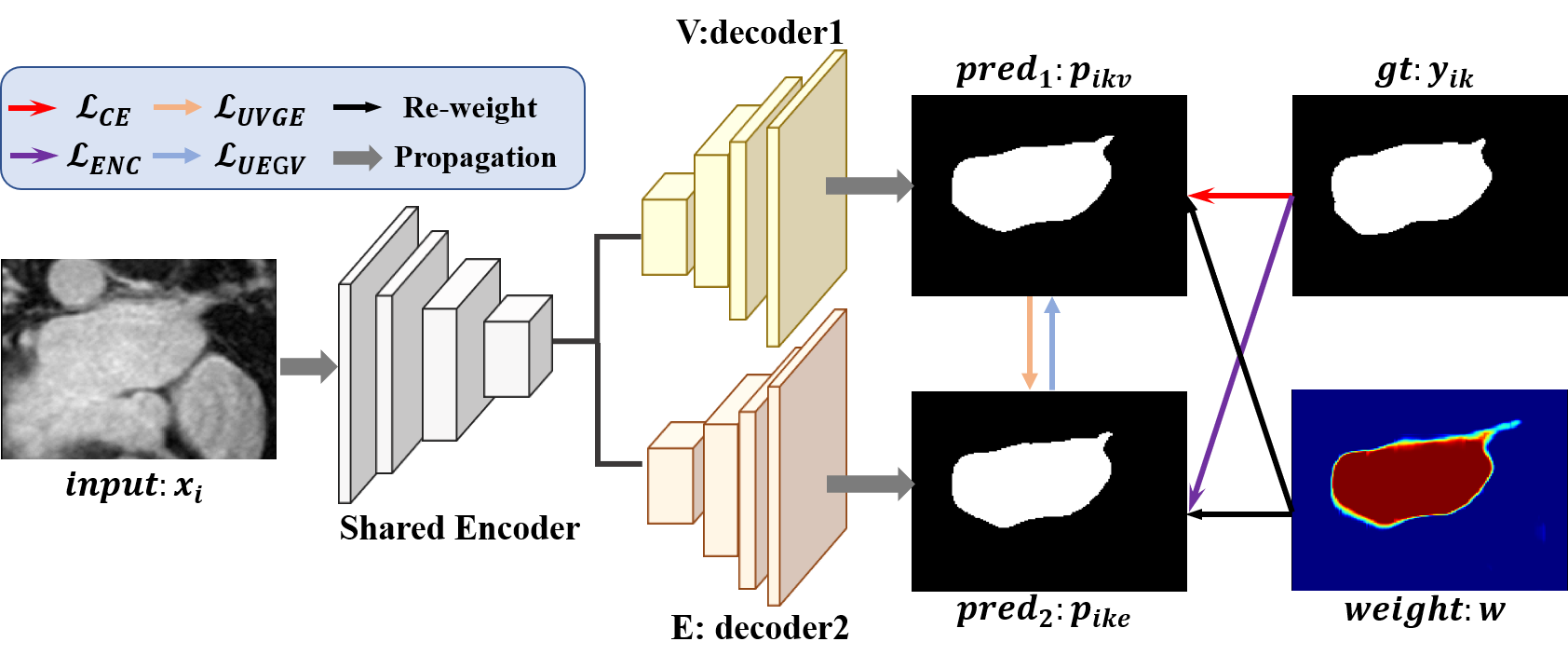}
	\caption{Pipeline of the proposed DC-Net.} \label{pipeline}
\end{figure}

\noindent\textbf{Vanilla Classifier.} This classifier is optimized by a common practice of the cross-entropy loss between the predicted probabilities $p_{ik}$ and the true one-hot label $y_{ik}$ as:
\begin{equation}
\mathcal{L}_{CE}=\sum_{k=1}^K-y_{ik} \log \left(p_{ik}\right)
\end{equation}
where $v$ means the segmentation probabilities are generated from vanilla classifier. $k$ is the class index and $K$ is the number of classes. $i$ denotes the pixel index.

\noindent\textbf{Evidential Non-Negative Classifier.}
This classifier is based on Evidential Deep Learning (EDL), which generalizes Bayesian theory to subjective probability \cite{jsang2016subjective} based on Dempster-Shafer Evidence Theory (DST) \cite{dempster1968generalization}. 
The belief distribution \cite{sensoy2018evidential} of DST in the discerning framework is represented as a Dirichlet distribution which is parameterized by $\alpha_{ik}=\left\{\alpha_{i1},\alpha_{i2},\cdots,\alpha_{iK}\right\}$.  
In semi-supervised segmentation, we define an evidence-based non-negative classifier which collects the predicted evidence $e_i \in \mathbbm{R}^K$ for different categories. The predicted evidence corresponds to the Dirichlet distribution with parameter $\alpha_{ik}= e_{ik}+1$. Then, the Dirichlet strength $S_i$, the predicted probability of pixel $p_{ik}$, and the overall uncertainty $u_i$ are formulated as:  
\begin{equation}
S_i=\sum_{k=1}^K \alpha_{ik}, \quad p_{ik}=\frac{\alpha_{ik}}{S_i}, \quad u_i=\frac{K}{S_i}
\end{equation}
The prediction confidence of each pixel $w_{i}$ is expressed as: 
\begin{equation}
w_i = 1- u_i
\end{equation}
where $w_{i} \geq 0$ and  $u_i \geq 0$.
Then, we optimize the Bayes Risk of $\mathcal{L}_{CE}$ with the Dirichlet distribution for the evidential classifier training, denoted as $\mathcal{L}_{ECE}$:
\begin{equation}
\mathcal{L}_{ECE}=\int\left[\sum_{k=1}^K-y_{i k} \log \left(p_{i k}\right)\right] \frac{1}{\mathcal{B}\left(\alpha_i\right)} \prod_{k=1}^K p_{i k}^{\alpha_{ik}-1} d p_i=\sum_{k=1}^K y_{i k}\left(\psi\left(S_i\right)-\psi\left(\alpha_{i k}\right)\right)
\end{equation}
where $\mathcal{B}$ is the multinomial beta function.
 $\psi\left( \cdot \right)$ denotes the digamma function.
A KL divergence loss function is introduced to further diminish the contribution of parameters associated with incorrect categories by reducing their evidence to 0 using Eq. \ref{KLeq}.
\begin{equation}
 \begin{aligned}
 \label{KLeq}
 \mathcal{L}_{{KL}}&=   {KL}\left[D\left(p_{ik} \mid \tilde{{\alpha}}_{ik}\right) \| D\left({p}_{ik} \mid \langle 1, \cdots, 1\rangle \right)\right]\\&=\log \left(\frac{\Gamma\left(\sum_{k=1}^K \widetilde{\alpha}_{i k}\right)}{\Gamma(K) \prod_{k=1}^K \Gamma\left(\widetilde{\alpha}_{i k}\right)}\right) 
 +\sum_{k=1}^K\left(\widetilde{\alpha}_{i k}-1\right)\left[\psi\left(\widetilde{\alpha}_{i k}\right)-\psi\left(\sum_{k=1}^K \widetilde{\alpha}_{i k}\right)\right]
 \end{aligned}
\end{equation}
where $\Gamma\left( \cdot \right)$ represents the gamma function. $\widetilde{\alpha}_{i k} = y_{ik}+\left({1}-{y}_{ik}\right) \bigodot \alpha_{ik}$ represents
the Dirichlet parameters after removal of the true evidence from the predicted parameters. 
The overall loss of the evidential non-negative classifier is described as:
\begin{equation}
\mathcal{L}_{ENC} = \mathcal{L}_{ECE} +\lambda_{kl} \mathcal{L}_{KL}
\end{equation}
$\mathcal{L}_{KL}$ is gradually incorporated into the training process with the control of an annealing coefficient $\lambda_{kl} = \min\left(1,t/150\right)$ for stable learning. $t$ denotes the current training iteration.
\subsection{Cross-Classifier Consistency Learning}
 
In order to leverage unlabeled data to improve the semi-supervised segmentation performance,
We propose a cross-classifier consistency loss that leverages complementary advantages of different opinions to improve segmentation performance. The prediction probability of one classifier is served as supervision information for another classifier. As shown in Fig.\ref{pipeline}, first, we use the predicted probability $p_{ike}$ from the evidential classifier to guide the output probability of the vanilla classifier $p_{ikv}$. We select $\mathcal{L}_1$ loss as the consistency constraint.
Further,  the uncertainty information generated by evidential classifier acts as a weight for $\mathcal{L}_{EGV}$. Since the high uncertainty regions typically relate to the noisy predictions in pseudo labels, which may cause error accumulation and impede the training process. If the model prediction has a high uncertainty degree $u_i$, then a smaller weight $w_i$ is applied to that prediction when calculating the loss.  The uncertainty weighted loss  $\mathcal{L}_{UEGV}$ can alleviate the negative impact of noise areas, leading to more robust and more trustworthy semi-supervised training.
\begin{equation}
\mathcal{L}_{UEGV} = \frac{1}{K}\sum_{k=1}^{K} w_i \cdot |p_{ike}-p_{ikv}|
\end{equation}
Similarly, we employ the predicted probability of the vanilla classifier to guide the training process of the evidential classifier. The uncertainty-weighted loss function $\mathcal{L}_{UVGE}$ is formulated as:
\begin{equation}
\mathcal{L}_{UVGE} = \frac{1}{K}\sum_{k=1}^{K} w_i \cdot |p_{ikv}-p_{ike}|
\end{equation}
Thus, the loss function for corss-classifier consistency  $\mathcal{L}_{CON}$ is defined as:
\begin{equation}
\mathcal{L}_{CON} = \mathcal{L}_{UEGV} + \mathcal{L}_{UVGE}
\end{equation}
\subsubsection{Total Loss of DC-Net}
The total loss of DC-Net consists of two parts: the supervised loss for labeled data $\mathcal{L}_{SEG}$ and the consistency loss $\mathcal{L}_{CON}$ for all data. $\mathcal{L}_{SEG}$ is only employed for labeled data, defined as:
\begin{equation}
\mathcal{L}_{SEG} = \mathcal{L}_{CE}(p_{ikv},y_{ik}) + \mathcal{L}_{ENC}(\alpha_{ik},y_{ik})
\end{equation}
The loss function $\mathcal{L}_{DCNet}$ of DC-Net is defined as:
\begin{equation}
\mathcal{L}_{DCNet} = \sum_{x_l \in \mathcal{D}_l}\mathcal{L}_{SEG} + \sum_{{x \in \mathcal{D}_l \cup \mathcal{D}_u}}\lambda_{c}\mathcal{L}_{CON}
\end{equation}
where $\lambda_{c}(t) = 0.1 \cdot e^{-5\left(1-\frac{t}{40}\right)^2}$ is a time-varying Gaussian ramp-up function to adjust the ratio between the supervised loss and unsupervised consistency loss.
\section{Experiment and Results}
\subsection{Data and Implementation Details}
We evaluate the proposed DC-Net on LA  dataset \cite{xiong2021global} and Pancreas-CT datasets \cite{clark2013cancer}. LA dataset is the benchmark in the 2018 atrial segmentation challenge, which contains 100 gadolinium-enhanced magnetic resonance imaging (MRI) scans with an isotropic resolution of $0.625 \times 0.625 ×\times0.625$ mm. 
We employ a fixed split of 80 training scans and 20 test scans. 
Pancreas-CT dataset includes 82 abdominal CT scans, which are acquired from the National Institutes of Health Clinical Center3. The resolution of axial slices is $512 \times 512$. the thickness ranges from 1.5 to 2.5 mm. We re-sample the data to have an isotropic resolution of $1 \times 1 ×\times1$ mm.
We employ a  fixed division of 62 training scans and 20 test scans. For pre-processing, we crop 3D samples from the original data to preserve the target region and normalize it to zero mean and unit variance as Wu et al \cite{wu2021semi}. Then, we randomly crop 3D patches with the size of $112 \times 112 \times80$ on LA dataset and $96\times96\times96$ on Pancreas-CT dataset, respectively. 
The batch size is 8, including 4 labeled patches and 4 unlabeled patches. SGD optimizer is employed with an initial learning rate of 0.1 and a poly learning rate decay policy.
The number of training iterations is 30000.
In the testing stage, we adopt a sliding window strategy with the fixed step size of 18 $\times$ 18 $\times$ 4 and 16 $\times$ 16 $\times$ 16) on LA dataset and Pancreas-CT dataset, respectively. 
We use four evaluation metrics to evaluate the performance, including Dice, Jaccard,
the average surface distance (ASD), and the 95\% Hausdorff Distance (95HD). We also report the number of model parameters (Para.) and the multiplicative-accumulative operations (MACs).   
\subsection{Comparisons with Other Methods}
To verify the effectiveness of the proposed method, we compare it with fully supervised (100\%) and limited supervised (20\%) baselines and relevant state-of-the-art methods on LA and Pancreas-CT datasets.

\begin{figure}[h]
	\includegraphics[width=\textwidth]{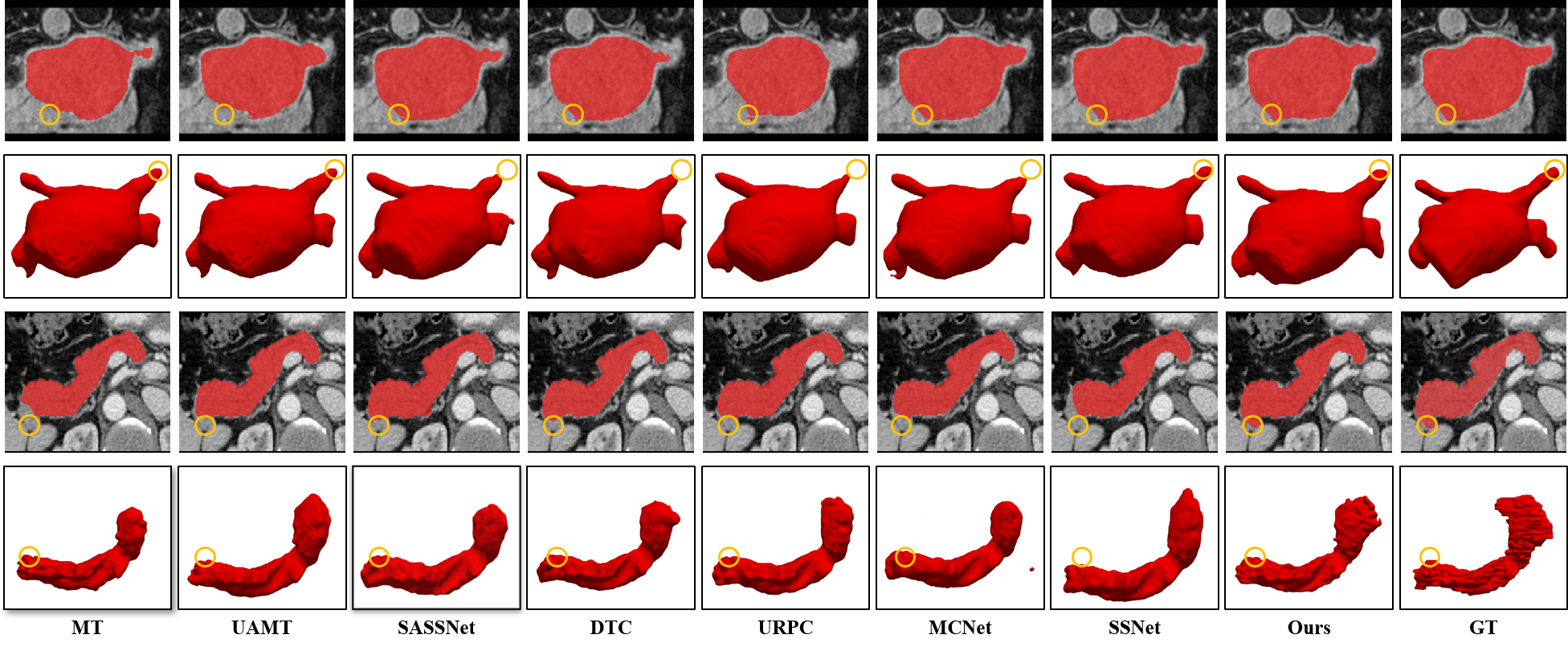}
	\caption{Visual comparision with other methods on LA dataset and Pancreas dataset in 2D view and 3D view.} \label{segres}
\end{figure}

\noindent\textbf{LA Dataset.}
Table \ref{tab1} shows the quantitative results of LA dataset. The first two rows give the results achieved by the fully-supervised V-Net using 20\% labeled data and all labeled data. The proposed DC-Net achieves comparable results, e.g., 90.89\% vs 92.27\% of Dice, compared with the fully supervised baseline by introducing the cross-classifier consistency. In addition, DC-Net gains a 4.86\% and 6.98\% improvement of Dice and Jaccard compared with the limited supervised baseline. In the same setting, DC-Net outperforms other cutting-edge semi-supervised methods on four evaluation metrics. All these quantitative results verify the superiority of our method. In addition, we visualize several 2D/3D segmentation examples in the first two rows of Fig. \ref{segres}. Our method generates consistent boundaries and complete structures as the yellow circles indicated. 
\begin{table}[h]
	\caption{Comparison with other methods on LA dataset.}\label{tab1}
	\centering
	\resizebox{\linewidth}{!}{
	\begin{tabular}{@{}lcccccccc@{}}
		\toprule
		\multirow{2}{*}{Method} & \multicolumn{2}{c}{Scans used} & \multicolumn{4}{c}{Metrics}                           & \multicolumn{2}{c}{Complexity} \\ \cmidrule(r){2-3} \cmidrule(r){4-7} \cmidrule(r){8-9}
		& Labeled      & Unlabeled      & Dice($\%$)$\uparrow$ & Jaccard(\%)$\uparrow$ & ASD$\downarrow$  & 95HD$\downarrow$  & Para.(M)       & MACs(G)       \\ \midrule
		V-Net                   & 16 (20\%)          & 0              & 86.03    & 76.08        & 2.45       & 10.55        & 9.44              & 47.02              \\
		V-Net                   & 80 (100\%)         & 0              & 92.27     & 85.69        & 1.30        & 5.25         & 9.44               &47.02               \\ \midrule
		MT \cite{tarvainen2017mean}                     & 16           & 64             & 87.28     & 77.93        & 2.49        & 11.17        & 9.44           & 47.02         \\
		UA-MT \cite{yu2019uncertainty}                  & 16           & 64             & 89.94     & 81.83        & 1.75        & 7.06         & 9.44           & 47.02         \\
		SASSNet \cite{li2020shape}                 & 16           & 64             &88.94           &80.32              &1.95             &9.17              & 9.44               &47.05               \\
		DTC \cite{luo2021semi}                   & 16           & 64             & 88.47     & 79.63        & 1.75        & 9.54         &9.44                &47.05               \\
		URPC \cite{luo2021efficient}                    & 16           & 64             & 90.46     & 82.67        &1.60             &6.77             &5.88              &69.43               \\
		MC-Net \cite{wu2021semi}                 & 16           & 64             & 90.20          &82.27              &1.72             &7.33              &12.35                &95.15               \\
		SS-Net \cite{wu2022exploring}                 & 16           & 64             & 88.77     & 80.09        & 1.74        & 7.70         &9.46                &47.17               \\
		DC-Net (Ours)           & 16           & 64             & \textbf{90.89}     & \textbf{83.06}        & \textbf{1.56}        & \textbf{6.76}         & 12.35               &95.15               \\ \bottomrule
	\end{tabular}}
\end{table}

\noindent\textbf{Pancreas-CT Dataset.}
Table \textcolor{red}{3} shows the quantitative results of Pancreas-CT dataset in supplementary material. Compared with the limited supervised baseline, DC-Net improves Dice, Jaccard, and 95HD by 7.21\%, 8.53\%, and 6.83, respectively. 
The Dice of DC-Net approaches the fully-supervised baseline (80.36\% vs 82.99\%). DC-Net outperforms all other methods on four metrics, demonstrating that DC-Net effectively exploits unlabeled data by cross-classifier consistency learning. From the \engordnumber{3} and  \engordnumber{4} row of Fig. \ref{segres}, we can observe that DC-Net achieves anatomically plausible segmentation results with more consistent boundaries compared to other methods.

\begin{table}[h]
	\caption{Ablation study of each design.}\label{tab3}
	\centering
	\begin{tabular}{@{}cccccccccc@{}}
		\toprule
		\multicolumn{2}{c}{Scans used} & \multicolumn{4}{c}{Loss}                                                                                       & \multicolumn{4}{c}{Metrics} \\ \cmidrule(r){1-2} \cmidrule(r){3-6} \cmidrule(r){7-10}
		Labeled       & Unlabeled      & $\mathcal{L}_{CE}$ & $\mathcal{L}_{ENC}$ &$ \mathcal{L}_{UEGV} $  &$\mathcal{L}_{UVGE}$  & Dice($\%$)$\uparrow$     &Jaccard(\%)$\uparrow$&ASD$\downarrow$      &95HD$\downarrow$      \\  \midrule
		16            & 0              & $\checkmark$                                     &-                                       &-                              &-  &86.38       &76.54       &2.33      &11.50      \\
		16            & 0              & $\checkmark$                                      &$\checkmark$                                        &-                             &-  &87.55       &78.29       &2.08      &8.97      \\
		16            & 64             & $\checkmark$                                     & $\checkmark$                                      &$\checkmark$                              &-  &89.63      &81.41       &1.93      &8.34    \\
		16            & 64             & $\checkmark$                                     & $\checkmark$                                      &-                             &$\checkmark$  &89.62       &81.38       &1.83      &7.71      \\
		16            & 64             & $\checkmark$                                     &$\checkmark$                                       &$\checkmark$                              &$\checkmark$  & \textbf{90.89}  &\textbf{83.06} &\textbf{1.56} &\textbf{6.76}  \\ \bottomrule
	\end{tabular}
\end{table}
\begin{figure}[h]
	\centering
	\includegraphics[width=0.75\textwidth]{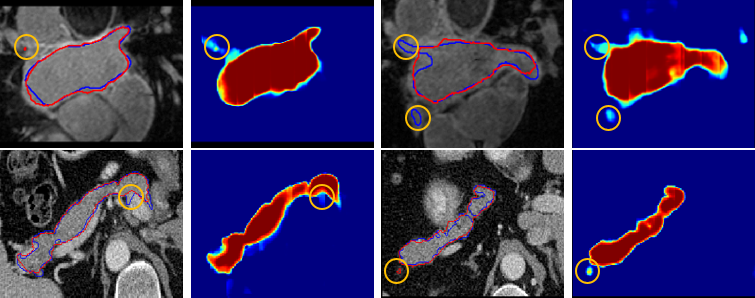}
	\caption{Visualization of the pseudo label and weight map $w$ generated by the evidential classifier: the red contour denotes the pseudo label and the blue contour denotes the ground truth. The heat map is the weighting in $\mathcal{L}_{UEGV}$ and $\mathcal{L}_{UVGE}$.} \label{uncertain}
\end{figure}
\subsection{Ablation Study}
We conduct the ablation study to validate the effectiveness of the critical designs of DC-Net on the LA dataset with 20\% labeled ratio. We gradually activate each loss component of the critical design to evaluate the model performance.

\noindent{\textbf{Effectiveness of Dual Classifiers.}}
As shown in Table. \ref{tab3}, the first setting only applies $\mathcal{L}_{CE}$, and the segmentation predictions are generated from the vanilla classifier alone. Then, the second setting introduces $\mathcal{L}_{ENC}$ additionally, from which the results are achieved by fusing the output of two classifiers. It becomes evident that the fusion leads to performance enhancements, highlighting the importance of introducing the evidential classifier.

\noindent{\textbf{Effectiveness of Cross-classifier Supervision.}}
Other settings (\engordnumber{3} to \engordnumber{5} row) in Table \ref{tab3} summarize the performance of cross-classifier consistency learning. When adding the cross-classifier guidance $\mathcal{L}_{UEGV}$ or $\mathcal{L}_{UVGE}$ (\engordnumber{3} and \engordnumber{4} row), performance evaluated by Dice increases 2.08\% and 2.07\%, respectively. In addition, we visualize the weight map $w$ in Fig. \ref{uncertain}. It’s obvious that the visualized weight maps show low responses (circulated in yellow) where the pseudo labels are inconsistent with ground truth, meaning that our method effectively adjusts the loss weight and mitigates the influences of erroneous supervision signals.

\section{Conclusion}
This paper proposes a cross-supervised learning framework based on dual classifiers, namely DC-Net, that guarantees a level of disagreement through their complementarity. The integration of evidential uncertainty into consistency learning reduces the negative impact of pseudo label noise.
Extensive experiments conducted on two public medical image datasets demonstrate the superior segmentation performance of DC-Net compared to other methods. In the future, we will 
investigate the interpretability of the evidential  classifier, which could provide valuable insights into the decision-fusion process of DC-Net .
%
\clearpage
\subsubsection{Supplementary Materials}
\begin{figure}[h]
	\centering
	\includegraphics[width=0.9\textwidth]{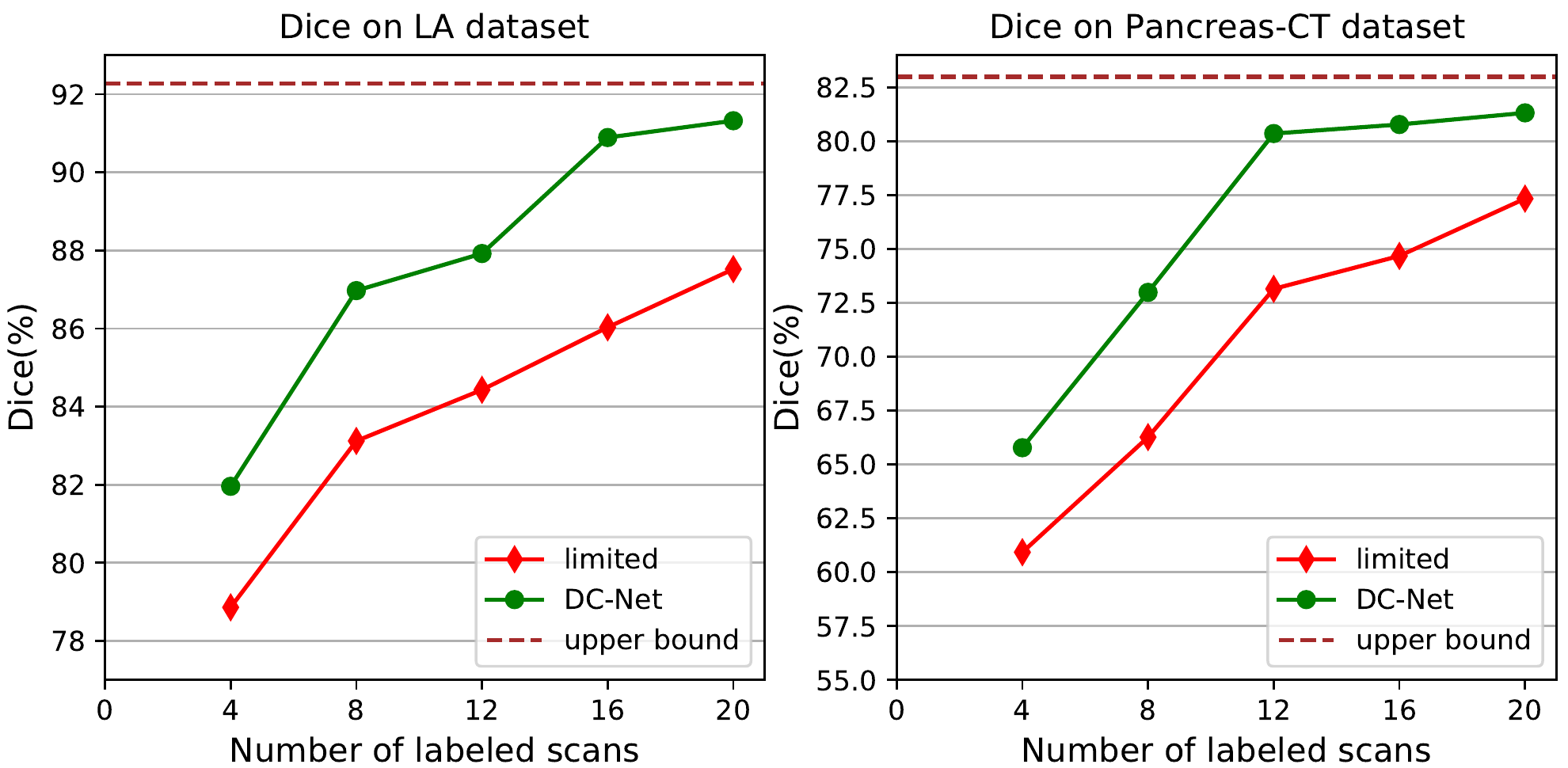}
	\caption{Segmentation results under different data splits: ``limited'' represent the limited supervised baseline. The upper bound performance is obtained by training with 100\% labeled data.} \label{ratioper}
\end{figure}

In this section, we report the result of comparative experiments on Pancreas-CT test set in Table \ref{PancreasCT}. Moreover, we report the results of semi-supervised DC-Net when using a different number of labeled scans in Table \ref{differentLA} and Table \ref{differentPanc} on LA dataset and Pancreas-CT dataset, respectively. Fig. \ref{ratioper} gives the change of Dice under different splits on both datasets. It can be observed that DC-Net yields consistent significant improvements compared with the limited supervised baseline which only uses the labeled data. DC-Net achieves comparable performance to the fully-supervised baseline with 20 labeled scans. (Dice: 91.32\% vs 92.27\% on LA, 81.32\% vs 82.99\% on Pancreas-CT). Our code will be released on GitHub after the review of this paper.
\begin{table}[h]
	\caption{Comparison with other methods on Pancreas-CT dataset.}
	\centering
	\resizebox{\linewidth}{!}{
		\begin{tabular}{@{}lcccccccc@{}}
			\toprule
			\multirow{2}{*}{Method} & \multicolumn{2}{c}{Scans used} & \multicolumn{4}{c}{Metrics}                           & \multicolumn{2}{c}{Complexity} \\ \cmidrule(r){2-3} \cmidrule(r){4-7} \cmidrule(r){8-9}
			& Labeled      & Unlabeled      & Dice($\%$)$\uparrow$ & Jaccard(\%)$\uparrow$ & ASD$\downarrow$  & 95HD$\downarrow$  & Para.(M)       & MACs(G)       \\ \midrule
			V-Net                   & 12 (20\%)          & 0              & 73.15     & 59.30        & 1.83        & 14.49        & 9.44              & 41.45              \\
			V-Net                   & 62 (100\%)          & 0              & 82.99     & 71.27       & 1.15        & 4.67         & 9.44               &41.45              \\ \midrule
			MT \cite{tarvainen2017mean}                     & 12           & 50             & 78.42     & 65.16        & 1.77        & 9.42        & 9.44           & 41.45       \\
			UA-MT \cite{yu2019uncertainty}                  & 12           & 50             & 78.49     &65.59        & 1.80        & 8.75         & 9.44           & 41.45         \\
			SASSNet \cite{li2020shape}                 & 12           &50             &79.34       &66.89              &1.40             &8.92     
			& 9.44               &41.48               \\
			DTC \cite{luo2021semi}                   & 12           & 50             & 77.69     & 64.42       & 1.51        & 8.57       &9.44                &41.48               \\
			URPC \cite{luo2021efficient}                    & 12           & 50             & 
			78.53    & 65.60        &{1.35}             &8.59             &5.88              &61.21               \\
			MC-Net \cite{wu2021semi}                 & 12           & 50             & 77.04          &63.53              &2.12             &8.44              &12.35                &83.88             \\
			SS-Net \cite{wu2022exploring}                 & 12         & 50             & 76.65     & 63.79        & 1.60        & 10.50        &9.46                &41.51               \\
			DC-Net (Ours)           & 12           & 50             & \textbf{80.36}     & \textbf{67.83}        & \textbf{1.30}        & \textbf{7.66}         & 12.35               &83.88               \\ \bottomrule 
	\end{tabular}}\label{PancreasCT}
\end{table}
\begin{table}[h]
	\caption{Segmentation results of DC-Net using different labeled scans compared with the limited supervised baseline on LA dataset.}
	\centering
	\begin{tabular}{@{}lcccccc@{}}
		\toprule
		\multirow{2}{*}{Method} & \multicolumn{2}{c}{Scan used} & \multicolumn{4}{c}{Metrics}                           \\ \cmidrule(r){2-3} \cmidrule(r){4-7} 
		& Labeled      & Unlabeled      & Dice(\%)$\uparrow$ & Jaccard(\%)$\uparrow$ & ASD$\downarrow$ & 95HD$\downarrow$ \\ \midrule
		V-Net                   & 80           & 0              & \textbf{92.27}     & \textbf{85.69}        & \textbf{1.30}        & \textbf{5.25}         \\ \midrule
		V-Net                   & 4            & 0              & 78.86     & 66.49        & 4.15        & 16.13        \\
		DC-Net                  & 4            & 76             & 81.96     & 70.17        & 3.84        & 14.67        \\
		V-Net                   & 8            & 0              & 83.12     & 72.66        & 2.98        & 10.94        \\
		DC-Net                  & 8            & 72             & 86.97     & 77.21        & 2.72        & 10.79        \\
		V-Net                   & 12           & 0              & 84.43     & 73.54        & 2.41        & 12.58        \\
		DC-Net                  & 12           & 68             & 87.92     & 78.85        & 2.10        & 8.52         \\
		V-Net                   & 16           & 0              & 86.03     & 76.08        & 2.45        & 10.55        \\
		DC-Net                  & 16           & 64             & 90.89     & 83.06        & 1.56        & 6.76         \\
		V-Net                   & 20           & 0              & 87.66     & 78.24        & 2.34        & 9.58         \\
		DC-Net                  & 20           & 60             & \textbf{91.32}     & \textbf{83.95}        & \textbf{1.51}        & \textbf{6.53}         \\ \bottomrule
	\end{tabular}\label{differentLA}
\end{table}

\begin{table}[!h]
	\caption{Segmentation results of DC-Net using different labeled scans compared with the limited supervised baseline on Pancreas-CT dataset.}
	\centering
	\begin{tabular}{@{}lcccccc@{}}
		\toprule
		\multirow{2}{*}{Method} & \multicolumn{2}{c}{Scan used} & \multicolumn{4}{c}{Metrics}                           \\ \cmidrule(l){2-7} 
		& Labeled      & Unlabeled      & Dice(\%)$\uparrow$ & Jaccard(\%)$\uparrow$ & ASD$\downarrow$  & 95HD$\downarrow$  \\ \midrule
		V-Net                   & 62           & 0              & \textbf{82.99}     & \textbf{71.27}        & \textbf{1.15}        &\textbf{4.67}         \\ \midrule
		V-Net                   & 4            & 0              & 60.93     & 45.79        & 2.08        & 25.89        \\
		DC-Net                  & 4            & 58             & 65.78     & 51.25        & 2.01        & 16.67        \\
		V-Net                   & 8            & 0              & 66.27     & 52.22        & 3.76        & 16.30        \\
		DC-Net                  & 8            & 54             & 72.99     & 58.78        & 1.91        & 16.92        \\
		V-Net                   & 12           & 0              & 73.15     & 59.30        & 1.83        & 14.49        \\
		DC-Net                  & 12           & 50             & 80.36     & 67.83        & 1.30        & 7.66         \\
		V-Net                   & 16           & 0              & 74.68     & 60.03        & 1.34        & 7.89         \\
		DC-Net                  & 16           & 46             & 80.78     & 68.12        & 1.26        & 6.76         \\
		V-Net                   & 20           & 0              & 77.43     & 64.12        & 1.60        & 9.79         \\
		DC-Net                  & 20           & 42             & \textbf{81.32}     & \textbf{68.45}        & \textbf{1.20}        & \textbf{6.53}         \\ \bottomrule
	\end{tabular}\label{differentPanc}
\end{table}
%
%
\clearpage
\bibliographystyle{splncs04_unsort}
\bibliography{refs}

\begin{thebibliography}{10}
\providecommand{\url}[1]{\texttt{#1}}
\providecommand{\urlprefix}{URL }
\providecommand{\doi}[1]{https://doi.org/#1}

\bibitem{mittal2019semi}
Mittal, S., Tatarchenko, M., Brox, T.: Semi-supervised semantic segmentation
  with high-and low-level consistency. IEEE Transactions on Pattern Analysis
  and Machine Intelligence  \textbf{43}(4),  1369--1379 (2019)

\bibitem{berthelot2019mixmatch}
Berthelot, D., Carlini, N., Goodfellow, I., Papernot, N., Oliver, A., Raffel,
  C.A.: Mixmatch: A holistic approach to semi-supervised learning. Advances in
  Neural Information Processing Systems  \textbf{32} (2019)

\bibitem{li2021dual}
Li, Y., Luo, L., Lin, H., Chen, H., Heng, P.A.: Dual-consistency
  semi-supervised learning with uncertainty quantification for covid-19 lesion
  segmentation from ct images. In: International Conference on Medical Image
  Computing and Computer-Assisted Intervention. pp. 199--209. Springer (2021)

\bibitem{yang2022st++}
Yang, L., Zhuo, W., Qi, L., Shi, Y., Gao, Y.: St++: Make self-training work
  better for semi-supervised semantic segmentation. In: Proceedings of the
  IEEE/CVF Conference on Computer Vision and Pattern Recognition. pp.
  4268--4277 (2022)

\bibitem{ouali2020semi}
Ouali, Y., Hudelot, C., Tami, M.: Semi-supervised semantic segmentation with
  cross-consistency training. In: Proceedings of the IEEE/CVF Conference on
  Computer Vision and Pattern Recognition. pp. 12674--12684 (2020)

\bibitem{ke2020guided}
Ke, Z., Qiu, D., Li, K., Yan, Q., Lau, R.W.: Guided collaborative training for
  pixel-wise semi-supervised learning. In: European Conference on Computer
  Vision. pp. 429--445. Springer (2020)

\bibitem{lai2021semi}
Lai, X., Tian, Z., Jiang, L., Liu, S., Zhao, H., Wang, L., Jia, J.:
  Semi-supervised semantic segmentation with directional context-aware
  consistency. In: Proceedings of the IEEE/CVF Conference on Computer Vision
  and Pattern Recognition. pp. 1205--1214 (2021)

\bibitem{wu2021semi}
Wu, Y., Xu, M., Ge, Z., Cai, J., Zhang, L.: Semi-supervised left atrium
  segmentation with mutual consistency training. In: International Conference
  on Medical Image Computing and Computer-Assisted Intervention. pp. 297--306.
  Springer (2021)

\bibitem{luo2021semi}
Luo, X., Chen, J., Song, T., Wang, G.: Semi-supervised medical image
  segmentation through dual-task consistency. In: Proceedings of the AAAI
  Conference on Artificial Intelligence. vol.~35, pp. 8801--8809 (2021)

\bibitem{fan2022ucc}
Fan, J., Gao, B., Jin, H., Jiang, L.: Ucc: Uncertainty guided cross-head
  co-training for semi-supervised semantic segmentation. In: Proceedings of the
  IEEE/CVF Conference on Computer Vision and Pattern Recognition. pp.
  9947--9956 (2022)

\bibitem{sensoy2018evidential}
Sensoy, M., Kaplan, L., Kandemir, M.: Evidential deep learning to quantify
  classification uncertainty. Advances in neural information processing systems
   \textbf{31} (2018)

\bibitem{zou2022tbrats}
Zou, K., Yuan, X., Shen, X., Wang, M., Fu, H.: Tbrats: Trusted brain tumor
  segmentation. In: Medical Image Computing and Computer Assisted
  Intervention--MICCAI 2022: 25th International Conference, Singapore,
  September 18--22, 2022, Proceedings, Part VIII. pp. 503--513. Springer (2022)

\bibitem{li2022region}
Li, H., Nan, Y., Del~Ser, J., Yang, G.: Region-based evidential deep learning
  to quantify uncertainty and improve robustness of brain tumor segmentation.
  Neural Computing and Applications pp. 1--15 (2022)

\bibitem{yu2019uncertainty}
Yu, L., Wang, S., Li, X., Fu, C.W., Heng, P.A.: Uncertainty-aware
  self-ensembling model for semi-supervised 3d left atrium segmentation. In:
  International Conference on Medical Image Computing and Computer-Assisted
  Intervention. pp. 605--613. Springer (2019)

\bibitem{wang2021tripled}
Wang, K., Zhan, B., Zu, C., Wu, X., Zhou, J., Zhou, L., Wang, Y.:
  Tripled-uncertainty guided mean teacher model for semi-supervised medical
  image segmentation. In: Medical Image Computing and Computer Assisted
  Intervention--MICCAI 2021: 24th International Conference, Strasbourg, France,
  September 27--October 1, 2021, Proceedings, Part II 24. pp. 450--460.
  Springer (2021)

\bibitem{jsang2016subjective}
Jsang, A.: Subjective logic: A formalism for reasoning under uncertainty.
  Springer Verlag  (2016)

\bibitem{dempster1968generalization}
Dempster, A.P.: A generalization of bayesian inference. Journal of the Royal
  Statistical Society: Series B (Methodological)  \textbf{30}(2),  205--232
  (1968)

\bibitem{xiong2021global}
Xiong, Z., Xia, Q., Hu, Z., Huang, N., Bian, C., Zheng, Y., Vesal, S.,
  Ravikumar, N., Maier, A., Yang, X., et~al.: A global benchmark of algorithms
  for segmenting the left atrium from late gadolinium-enhanced cardiac magnetic
  resonance imaging. Medical Image Analysis  \textbf{67},  101832 (2021)

\bibitem{clark2013cancer}
Clark, K., Vendt, B., Smith, K., Freymann, J., Kirby, J., Koppel, P., Moore,
  S., Phillips, S., Maffitt, D., Pringle, M., et~al.: The cancer imaging
  archive (tcia): maintaining and operating a public information repository.
  Journal of Digital Imaging  \textbf{26}(6),  1045--1057 (2013)

\bibitem{tarvainen2017mean}
Tarvainen, A., Valpola, H.: Mean teachers are better role models:
  Weight-averaged consistency targets improve semi-supervised deep learning
  results. Advances in neural information processing systems  \textbf{30}
  (2017)

\bibitem{li2020shape}
Li, S., Zhang, C., He, X.: Shape-aware semi-supervised 3d semantic segmentation
  for medical images. In: Medical Image Computing and Computer Assisted
  Intervention--MICCAI 2020: 23rd International Conference, Lima, Peru, October
  4--8, 2020, Proceedings, Part I 23. pp. 552--561. Springer (2020)

\bibitem{luo2021efficient}
Luo, X., Liao, W., Chen, J., Song, T., Chen, Y., Zhang, S., Chen, N., Wang, G.,
  Zhang, S.: Efficient semi-supervised gross target volume of nasopharyngeal
  carcinoma segmentation via uncertainty rectified pyramid consistency. In:
  International Conference on Medical Image Computing and Computer-Assisted
  Intervention. pp. 318--329. Springer (2021)

\bibitem{wu2022exploring}
Wu, Y., Wu, Z., Wu, Q., Ge, Z., Cai, J.: Exploring smoothness and
  class-separation for semi-supervised medical image segmentation. In: Medical
  Image Computing and Computer Assisted Intervention--MICCAI 2022: 25th
  International Conference, Singapore, September 18--22, 2022, Proceedings,
  Part V. pp. 34--43. Springer (2022)

\end{thebibliography}

\end{document}